# A STEP TOWARDS THE APPLICABILITY OF ALGORITHMS BASED ON INVARIANT CAUSAL LEARNING ON OBSERVATIONAL DATA

**Borja Guerrero Santillan**
Universidad Politécnica de Madrid(UPM), ETSIINF, Spain
b.guerrero@alumnos.upm.es

March 30, 2023

## ABSTRACT

Machine learning can benefit from causal discovery for interpretation and from causal inference for generalization. In this line of research, a few invariant learning algorithms for out-of-distribution (OOD) generalization have been proposed by using multiple training environments to find invariant relationships. Some of them are focused on causal discovery as Invariant Causal Prediction (ICP), which finds causal parents of a variable of interest, and some directly provide a causal optimal predictor that generalizes well in OOD environments as Invariant Risk Minimization (IRM). This group of algorithms works under the assumption of multiple environments that represent different interventions in the causal inference context. Those environments are not normally available when working with observational data and real-world applications. Here we propose a method to generate them in an efficient way. We assess the performance of this unsupervised learning problem by implementing ICP on simulated data. We also show how to apply ICP efficiently integrated with our method for causal discovery. Finally, we proposed an improved version of our method in combination with ICP for datasets with multiple covariates where ICP and other causal discovery methods normally degrade in performance.

## 1 Introduction

Current machine learning approaches suffer from a fundamental problem. They are limited to pattern recognition and learning complex prediction rules by minimizing a training error. As a consequence machine learning systems rely on data biases based on possible spurious correlations without understanding cause and effect relations in the data. This leads to a lack of generalizability with a limited ability to predict in unseen contexts. Causation is then key to achieve out of distribution generalizability and building more robust systems. that can answer interventional questions. Another important limitation of the current machine learning approaches is explainability. For the same reason, not really understand the causes and effects of the data. While methods based on model interpretation like LIME [22] and SHAP [17] can help on this issue, they do not really answer how data behave, just how a model does. Again to overcome this limitation causation is key. Then, we have seen how important is to infer cause and effect relationships. Generally, it is a standard procedure to implement a randomized controlled trial, whereby a random assignment of one population is applied and intervention(change in a variable of interest) while another population is not. From this experiment, the data is collected and used to measure the cause and effect of that intervention. However, these kinds of experiments can be difficult or impossible to conduct or simply not worthy due to cost, time, and other limitations. In those cases, we work with observational data and apply quantitative methods to discover causal relations, which is known as causal discovery.

Causal discovery is a challenging field and it is often performed by domain knowledge rather than quantitative tools. However, if there is no previous knowledge of possible causal relations we need mathematical methods to be applied. Causal discovery is based on the theory of structural causal models(SCM) [3],[24],[19] and causal graphs [16], [9], [27], [23], [25]. A causal graph is a directed acyclic graph denoting the dependency between variables. A structural causal model also specifies the functional form of dependencies between variables. Between the computational methods based on graphical models, we can distinguish constraint-based and score-based methods versus the ones based on functional causal models. The PC and the Fast Causal Inference(FCI)[26] are constraint-based and work testing for conditional independence on a starting plausible causal graph. On the other hand, the Greedy Equivalence Search(GES) [5] aims to build the causal graph from zero checking if the causal relation to be added optimizes a defined score function. Also, we can find SCM algorithms based on non- gaussian models [25] and non-linear models [12],[21] which generally are able to distinguish between different Directed Acyclic Graphs but at the cost of making the additional assumptions of the SCM. However in general these methods have limitations since multiple causal structures can satisfy the same conditional independence. Therefore, often they can provide a set of plausible causal graphs which actually may not contain the true one.

More recently, a research wave has focused on the invariant property of causal relations for causal discovery and causal inference purposes. One of the first was the Invariant Causal Prediction (ICP) [20] which is a causal discovery algorithm based on considering all direct causes of a target variable of interest. This approach tries to exploit the well-known invariance property of causal relations as opposed to spurious correlations to infer causality. Specifically, the method hypothesis is that the conditional distribution of a target variable of interest given a set of direct causal predictors has to remain identical under different interventions. In this context, they define the concept of environments which are different datasets, each one representing a particular intervention. Then the algorithm iterates over multiple combinations of subsets of features to find the ones that are invariant as plausible causal parents of the target variable. Finally, the intersection of these sets of plausible causes, which are predictive in all environments, is then a subset of the true direct causes.

Invariant Risk minimization(IRM) [2] is another causal discovery algorithm based on the same idea of exploiting the invariant property of causal relation for inference. IRM also relies on the idea of training in multiple environments. However, IRM does not focus on retrieving the causal parents of the target in a causal graph. Rather, IRM aims to achieve out-of-distribution generalization for a predictive model in unseen environments. The IRM approach is based on optimizing a penalty function used to obtain a data representation, on which a model can perform optimally in all environments. Other related work has followed up with this concept and the ultimate goal of improving OOD generalization in different machine learning tasks [4], [7],[13], [1],[14]. Some of them are focused on specific cases such as non-linear ones [11] and robustness, especially in settings with high data shifts [10],[13], [18],[15].

In most of this work, the environment concept plays a central role. It is a general assumption to have given environments on the data. Following up with a realistic context and when working with observational data this is not normally the case and therefore it is not possible to apply these types of algorithms. Some recent work, specifically in the context of deep learning, has incorporated splitting methods to generate the environments [28]. Also, ISL[6] has been the first to incorporate unsupervised learning, in particular a K-means, into an invariant learning algorithm.

Here we proposed a method that is based on 'supervised clustering' to generate the environments which can be combined or not with ICP for causal discovery, and that can be combined with another algorithm that is based on the causal invariant property. We strictly interpret the assumptions made in the given environments of original work such as ICP and IRM. It is therefore an open door to implement these kinds of algorithms. It is suitable for observational data and real-world applications in general. Our method works well on mixed datasets with continuous, categorical, and binary features and also even when there are no high shifts in the data given, which can be often more challenging than the opposite case. We assess the performance of our method by combining it with ICP on multiple synthetic causal datasets with unknown environments so that we can check the real efficiency of the method.

## 2 Methodology

As we have seen, when working with observational data normally we do not know the environment needed to apply the invariant property. Hence we need a method to generate that environment for a given observational dataset.

### 2.1 Generating environments

In the context of causal discovery and ICP algorithms, we can define a set of inputs or covariates X and a target variable Y whose causal parents we want to study. Then, in the invariant causal algorithms, the main assumption that researchers take is that these environments represent different sources that could represent different interventions of a randomized experiment. Then these environments present by definition a change in the distribution of the covariates between them, which is known in the machine learning literature as covariate shift. Generally, this term refers to the shift between training and test set when in this context is between environments.

More strictly other researchers define these environments as presenting a change not only in the inputs but in the joint distribution between inputs and target i.e P(X, Y), which is generally addressed as dataset shifts. Since this assumption is necessary to apply ICP successfully it is our goal when generating environments. In this work, we propose a method based on a strict interpretation of this assumption that guarantees dataset shift between environments. This assumption can be written
as $P(X_i, Y)_{ej} \neq P(X_i, Y)_{ek}$ when $j \neq k$ and $\forall e \in E$. A loose interpretation would be to generate a covariate shift between environments. The second interpretation could be not optimal and therefore have a higher risk of not finding the existing causal relations.

The environments are in other words latent clusters that are unknown. Since we have addressed the necessity to generate a dataset shift between them we propose a supervised clustering approach, an unsupervised classification that uses a supervised method. That way we can manage to obtain clusters with different joint distributions between inputs and target variables of interest. However, not all supervised models can provide identifiable clusters. Here we proposed a decision tree as the algorithm that can meet both requirements. First, a decision tree can be fitted as a supervised algorithm and it maximizes the heterogeneity between the nodes generated by each split. In other words, it maximizes the variance between nodes taking into account the input and its relations with the output, which is precisely what we want to achieve. On the other hand, the splits are perfectly defined by the threshold on the features which are known as the splitting conditions, and therefore we can obtain the clusters where each final node constitutes an environment.

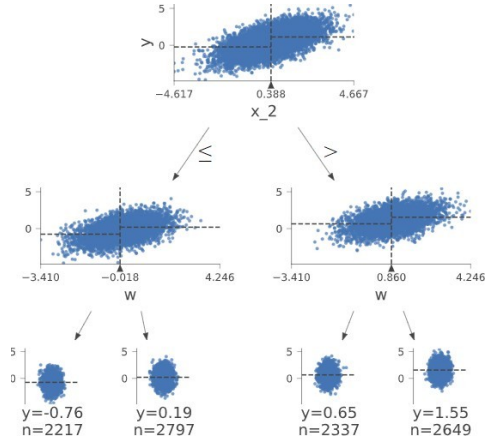

Figure 1: Decision Tree Clustering.

One important aspect is that each variable can meet the requirement. That is why the generation of the environment is conditional on each input variable and the target of interest. Hence, for a given fixed number E of environments, we will generate E environments for each $X_i$ in X conditioning on Y. That way we can guarantee testing causality for each covariate. Regarding the number of clusters, we prefer a medium number of them rather than a small one. Informally that is because it is a more robust result and also can a better guarantee than having by instance only two environments. On the other hand, we have not seen the special advantages of using a high number of environments.

### 2.2 Theoretical Framework

Once we have generated the environments we can exploit the property of causal invariance to perform causal discovery on observational data. We recall the invariance assumption, as [20] originally defined, and discuss the notion of identifiable causal predictors. Let's define E as the space of all possible interventions and therefore E contains all the
individual environments e. As stated above, we have variables $(X)_e$, $(Y)_e$ with a joint distribution that will in general depend on the environment $e \in E$.

**Proposition 1 of Invariant prediction**: *There exists a vector of coefficients $\gamma\star$ of dimension p, where p is the number of real causal parents of y, that satisfies for all* $e \in E$ ::

$$(Y)_e = \mu + (X)_e \gamma \star + (\epsilon)_e, \quad \epsilon_e \sim F_\epsilon, \quad \epsilon_e \perp\!\!\!\perp X_e \in S^\star \tag{1}$$

where $\epsilon_e$ is an error with mean zero, finite variance, and the same distribution $F\epsilon$ across all $e \in E$. Hence, the variables that have a direct causal effect on Y in a SEM form a set $S^\star$ for which Proposition 1 is satisfied. Therefore the first implication of the invariant prediction assumption is that given a subset of variables $S^\star$ that contain the real causal parents of Y its vector of coefficients is constant between environments. The second implication is that the residuals of the model must have equal mean across environments and, in a strict interpretation, equal variance too.

More formally, and from a probabilistic point of view, we can define the invariant causal property as equality between environments of the conditional probability of the target variable given the causal parent. Then the complete theoretical framework that we propose for causal discovery has to meet two conditions:

1. Inequality of joint distributions of target and input variables among environments, sometimes referred to as dataset shift.

$$P(X_i, Y)_{ej} \neq P(X_i, Y)_{ek}, \qquad j \neq k, \quad \forall e \in E \qquad (2)$$

where i = 1,2,...,N being N the number of covariates, and j, k = 1,2,...,K being K the number of environments.

2. Equality of conditional probability of the target variable given the causal parent.

$$P(X_i|Y)_{ej} = P(X_i|Y)_{ek}, \qquad j \neq k, \quad \forall e \in E \qquad (3)$$

where i = 1,2,...,N being N the number of covariates, and j, k = 1,2,...,K being K the number of environments.

Since our method generates environments associated with each variable individually to assure that condition 1 is met, we have to check also condition 2 individually for the corresponding environment e, and only when both conditions are met do we consider $X_i$ to be a causal parent of the target variable Y.

## 2.3 The algorithm

Having understood the theoretical framework we can define the steps of the algorithm. Given a dataset where we select a target variable to be studied Y and the rest of the covariates as X, the steps are:

1. Fit a decision tree on every $X_i$ and Y with a fixed number K of environments, as a classifier for binary or categorical target variables and as a regressor otherwise.
2. Keep the thresholds defined by the tree for each variable to form the environment. With this, we obtain K environments ($e_1i, e_2i, \ldots, e_ki$) for each $X_i$. We can perform a two-sample KS test to check the dataset shift between environments.
3. For all possible subsets of input variables:
   (a) Train a model using all environments, only the specific input features in that subset, and obtain the residuals. For the linear case, we fit a linear regression with estimated parameters $\beta(S)$ and residuals $R = Y - X\beta(S)^*$.
   (b) Test the null hypothesis that the mean of the residuals is equal for every environment using a two sample T-test and Bonferroni correction when combining all. For equal variance use an F-test and Bonferroni correction. Finally, combine both p-values using twice the minimum value and reject the subset if the final p-value is lower than a level of confidence α.
4. Find the features present in all the not rejected subsets as potential causal predictors. Only keep the variables $X_i$ that are candidates to causal parent on its related environments ($e_1i, e_2i, \ldots, e_ki$) as selected causal parent of the target variable Y.

For cases with multiple covariates, ICP can decrease in performance since the problem becomes more challenging due to its conservative nature. For these cases, we propose a different version of the same algorithm where the goal is to build a more robust procedure to find the candidates for causal parents. The differences are in steps:

3. Before generating all possible subsets of input variables we limit the size of inputs to N=5 and we generate all possible subsets. Now each of these subsets becomes our original inputs and we apply step 3 as in the original version of the algorithm.
4. Since we have performed multiple checks we propose a voting approach. It consists of counting the number of each possible candidate(following the same logic as before) and we use a new confidence interval to keep the final selected parent. For instance, if we choose α = 0.1 we only accept as a causal parent a candidate that has been selected as a candidate in at least 90 % of checks performed.

This second version has proved to achieve good results over the original version of the algorithm for cases with multiple covariates and more complex causal relations as we will see in the next section.

## 3 Numerical Results

In this section, we distinguish two experiments. Since we are solving an unsupervised learning problem when generating the environment we need a way to evaluate the method. For that purpose, we will use the environment to apply ICP following the algorithm described in the previous section. The implementation of ICP used in these experiments is an adapted version of the Python library ICPy [8]. Then, we apply the method to simulated data and measure the causal discovery capacity by the True Positive Rate (TPR) and the False Discovery Rate (FDR). As a benchmark of these metrics, we implement other traditional causal discovery algorithms including FCI y Lingam. The goal of this comparison is double, first, it allows us to test ICP´s power when finding causal relations and second and more importantly it can let us know if our algorithm is efficient when generating the environments.

In the first experiment, we generate 3 different datasets. Some of the covariates are randomly generated through a gaussian process or binomial process. The rest of the variables are generated through structural equations models (SEMs) and gaussian noise. The datasets go from simple to complex and cover different settings. In particular, the second dataset includes measuring confounding through variables that affect both cause and effect. The second dataset includes binary and categorical variables as well where our method is adapted to work well, both in the first part of the method and also the ICP algorithm.

Then we simulate each dataset 5 times and apply our method. Here we apply iteratively to every variable of each dataset. Therefore, each variable becomes the target variable of interest with the intention to try to rebuild the complete causal graph. This way we are testing the algorithm in a more challenging way and checking its reliability. Also, we can compare it easily with the other causal discovery algorithms. Finally, we measure the TPR and the FDR as the average of the 5 simulations to form a robust metric. Table 1 and Table 2 contain the TPR and the FDR of this experiment respectively.

|          | ICP | FCI | LINGAM |
|----------|-----|-----|--------|
| Dataset1 | 1.0 | 1.0 | 1.0    |
| Dataset2 | 1.0 | 1.0 | 1.0    |
| Dataset3 | 1.0 | 1.0 | 1.0    |

Table 1: True Positive Rates Comparison of Experiment 1

|          | ICP  | FCI  | LINGAM |
|----------|------|------|--------|
| Dataset1 | 0    | 0.08 | 0.16   |
| Dataset2 | 0    | 0    | 0.14   |
| Dataset3 | 0.04 | 0    | 0.20   |

Table 2: False Discovery Rates Comparison of Experiment 1.

Regarding the interpretation, the TPR gives a percentage of how many causal parents of the graph have been detected by the algorithm where 1 is the best situation and 0 is the worst. And the FDR describes how many times a variable has been selected as a causal parent and it is actually not, being 0 the best scenario and 1 the worst one. Looking at the results in table 1 we can see how ICP can compete with the other methods with True Positive Rates of 100% meaning that can find mostly all of the causal predictors of all the variables. On table 2 we also see results with the highest rate around 4% versus 8% for FCI and 20 % for LINGAM. Based on these results, we can claim that the generation of the environments is efficient to check for the invariant property of causal relations on the data.

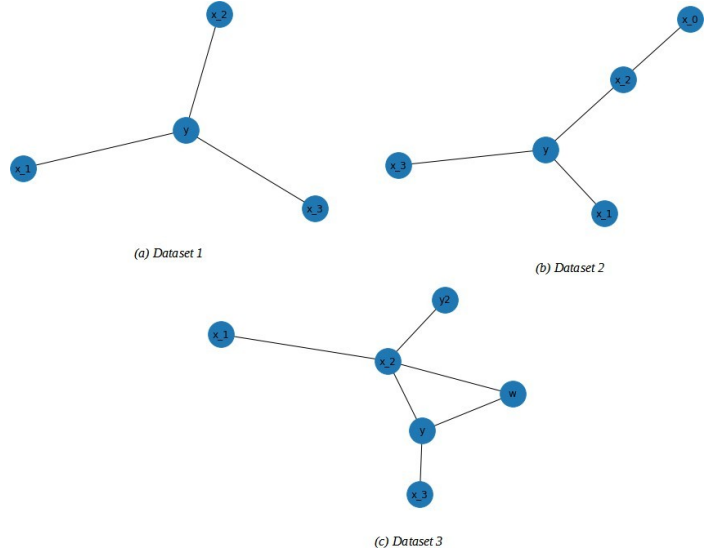

Figure 2: Causal Graph discovered by ICPv1 on Dataset 1, 2 and 3.

For the second experiment, we generate two new datasets. Following the same logic as before these datasets are more complex, they include more covariates and more causal relations. It is well known that when the number of covariates grows causal discovery algorithms decrease their performance. On the other hand, ICP is known to suffer from the same problem. Testing those limitations is precisely the goal of this second experiment. We will refer to the first method as ICPv1 and the second as ICPv2.

As a consequence of those limitations and the preliminary results, we have developed another version of our algorithm as we have explained already in section 3. This version of the algorithm is based on the same concepts and its main goal is to be more robust in datasets with multiple covariates. We include it in the comparison of this experiment and discuss the results of Tables 3 and 4.

|  | FCI | LINGAM | ICPv1 | ICPv2 |
| --- | --- | --- | --- | --- |
| Dataset5 | 0.73 | 1.0 | .73 | 1.0 |
| Dataset6 | 0.87 | 1.0 | 0.6 | 1.0 |

Table3: True Positive Rates Comparison of Experiment 2.

|  | FCI | LINGAM | ICPv1 | ICPv2 |
| --- | --- | --- | --- | --- |
| Dataset5 | 0 | 0.21 | 0.11 | 0.08 |
| Dataset6 | 0 | 0.38 | 0.1 | 0.21 |

Table4: False Discovery Rates Comparison of Experiment 2.

Looking at tables 3 and 4 we can see how the previous version of the algorithm fails to perform causal discovery in an efficient way in this case showing poor TPR and FDR. In a similar way, FCI and LINGAM performance highly decrease as well showing also poor rates. Finally, we can see how the new version proposed achieves good rates for these complex graphs.

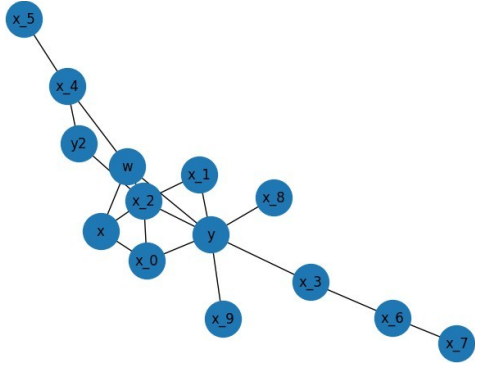

Figure 3: Causal Graph discovered by ICPv2 on Dataset 5.

## 4 Conclusion

Based on the results, we can claim that our method is suitable for generating environments efficiently on observational data. This opens a door to applying algorithms based on the invariant causal property, which is one of the main contributions of this paper and can be used to perform causal inference, causal discovery, or simply improved machine learning predictions(as IRM) on observational data. We also have seen how this can be implemented in combination with the ICP algorithm in a successful way to perform causal discovery on observational data. Finally, we have seen a new version of ICP in combination with our method suitable for datasets with multiple covariates and complex causal relations. This new version shows good TPR and FDR and has proved to outperform other causal discovery methods, hence it can become an alternative for causal discovery. Like other causal discovery algorithms, our method is not able to find the direction of the causal relation, which is a limitation and could be a goal for future work.

# A  Appendix

## A.1  Experimental settings for numerical results

The datasets are generated according to the criteria specified in the numerical results section. Specifically, the five datasets are simulated according to the following equations:

*Dataset1*

$X_1 = N(0.8, 1)$

$X_2 = N(0.5, 1)$

$Y = 0.7X_1 + 0.4X_2 + N(0, 1)$

$X_3 = 0.5Y + N(0, 1)$

*Dataset2*

$X_1 = Bin(6, 0.5)$

$X_2 = Bin(0.5, 1)$

$Y = 0.7X_1 + 0.4X_2 + N(0, 1)$

$X_3 = 0.5Y + N(0, 1)$

*Dataset3*

$X_1 = N(0.2, 0.5)$

$W = N(0.5, 1)$

$X_2 = 0.6W + 0.4X_1 + N(0, 1)$

$Y = 0.5X_2 + 0.5W + N(0, 1)$

$Y_2 = 0.7X_2 + N(0.2, 1)$

$X_3 = 0.3Y + N(0, 1)$

*Dataset4*

$X = N(0.2, 0.8)$

$X_1 = N(0, 1.1)$

$X_0 = 0.5X + N(0, 0.5)$

$W = X + N(0, 1)$

$X_2 = 0.5X_1 + 0.5W + 0.5X_0$

$Y = 0.5X_2 + 0.5W + N(0, 1)$

$Y_2 = 0.7X_2 + N(0.2, 1)$

$X_3 = 0.5Y + N(0, 1)$

$X_4 = W + N(0, 1)$

$X_5 = 0.8X_4 + N(0, 1)$

*Dataset5*

$X = N(0.2, 0.8)$

$X_1 = N(0, 1.1)$

$X_9 = N(0.4, 0.75)$

$X_0 = 0.5X + N(0, 0.5)$

$W = X + N(0, 1)$

$X_2 = 0.5X_1 + 0.5W + 0.5X_0$

$Y = 0.5X_2 + 0.5W + 0.5X_9 + N(0, 1)$

$Y_2 = 0.7X_2 + N(0.2, 1)$

$X_3 = 0.5Y + N(0, 1)$

$X_4 = W + N(0, 1)$

$X_5 = 0.8X_4 + N(0, 1)$

$X_6 = X_3 + N(0, 1)$

$X_7 = 0.1X_6 + N(0.2, 0.5)$